%% file: arxiv.tex
\documentclass{article}
\usepackage{preprint,times}
\input{math_commands.tex}

\usepackage{hyperref}
\usepackage{url}
\usepackage{amsmath,amssymb,amsthm}
\usepackage{booktabs}
\usepackage{placeins}
\usepackage{graphicx}
\usepackage{xcolor}
\usepackage{tikz}
\usepackage{algorithm}
\usepackage{algorithmic}
\usetikzlibrary{calc}

\renewcommand{\eqref}[1]{(\ref{#1})}

\newtheorem{theorem}{Theorem}

\newtheorem{proposition}{Proposition}
\newtheorem{lemma}{Lemma}
\newtheorem{corollary}{Corollary}
\newtheorem{definition}{Definition}

\title{Noise-Level Adjacency in Diffusion Training}

\author{T. Y. Tsui, Jiatao Gu, Lingjie Liu \\
University of Pennsylvania}

\begin{document}

\maketitle

\begin{abstract}
Diffusion models train one network across all noise levels, although their training objective is pointwise in the noise level. If training on each pair of adjacent levels were an independent problem, separate networks with the same total parameter count would be expected to match one shared network. But in practice, on images, sharing weights across levels gives better results, which hints at an implicit dependence on their order. An order fixes both which levels are adjacent and the direction in which they run. In this paper, we show that, of the two, the benefit of sharing rests on adjacency in the noise-level embedding. The fitting error of a shared network is bounded below by the distances between neighbouring posterior means, minus the change the network can make between neighbouring embeddings. We first distinguish adjacency from direction: on a Gaussian mixture and CIFAR-10, reversing the embeddings keeps every neighbour and leaves the fitting error unchanged, whereas shuffling them raises it. Second, we rule out sharing between distant levels as the source of the benefit: separate networks on contiguous ranges match the shared network at its width, and with the parameters fixed, breaking adjacency at the same places costs less than 40\% of a full shuffle. Dividing the levels at equal shares of these distances lowers the error of all but the smallest networks. We therefore show that the order of the noise levels enters a diffusion model through adjacency, with its direction being a design choice. Our code is available at https://github.com/TSUITUENYUE/Noise-Level-Adjacency-in-Diffusion-Training.
\end{abstract}

\section{Introduction}

Diffusion models were introduced through a physical process: a Markov chain from non-equilibrium thermodynamics destroys the data \citep{sohl2015deep}. Later work showed that much of this description is a design choice \citep{song2021ddim, karras2022edm, lipman2023flow}. However, across these formulations one network is trained at every noise level, and work that reads the levels as tasks of one network measures how they transfer to and conflict with one another \citep{go2023negative, hang2023minsnr}. We ask whether the order of the noise levels is also a design choice.

Each term of the training objective is evaluated at one noise level on samples drawn from the closed-form marginal \citep{ho2020ddpm}. The objective therefore separates across the levels and does not change under any relabelling of them. \citet{song2021ddim} used this separability to show that, when parameters are not shared across noise levels, the optimum does not depend on how the levels are weighted. The model learns one pointwise quantity at each noise level, and the order of the levels is fixed by hand in the sampler that composes them. Objectives that compare predictions at two noise levels, such as progressive distillation and consistency models, write an order into the objective itself \citep{salimans2022progressive, song2023cm, kim2024ctm}.

However, training does not treat each pair of noise levels as an independent problem. If it did, separate networks with the same parameters in total would match one shared network; on image generation, the shared network fits better. Sharing must therefore use a relation among the noise levels that the data carry and the objective does not couple, and it can reach that relation only through the parameters those noise levels share.

The answer matters for two choices that diffusion models currently make by hand. A network reads the noise level through an embedding chosen by convention, usually sinusoidal features of the level, and recent work asks whether it needs the level at all \citep{sun2025noise}. Some models divide the noise levels among expert networks at boundaries fixed by hand \citep{balaji2022ediff, podell2024sdxl, lee2024meme}. Knowing what sharing uses tells, from the data alone, what an embedding must preserve and which levels a network can hold together.

\begin{figure}[t]
\centering
\includegraphics[width=\textwidth]{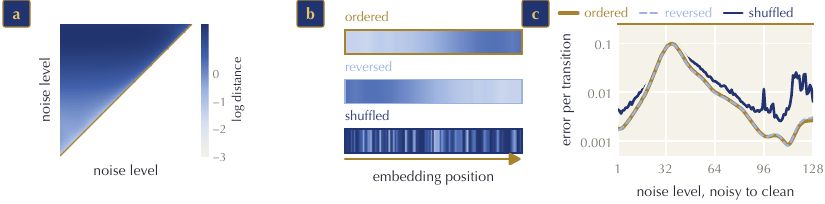}
\caption{Where the order enters. (a) Closed-form distances between posterior means of the Gaussian mixture on the pooled inputs of every level, from noisiest at bottom left to clean at top right (light: near; dark: far). (b) Distances between posterior means at consecutive embeddings under ordered, reversed and shuffled assignments, on the same scale. (c) Error by noise level for a width-256 network under those assignments, averaged over 5 seeds.}
\label{fig:orderenters}
\end{figure}

In this paper we show that this relation is adjacency. An order fixes both which levels are adjacent and the direction in which they run, and the benefit of sharing rests on which levels are neighbours in the noise-level embedding, with almost no dependence on their direction. A shared network tells the levels apart through the embedding (Figure~\ref{fig:orderenters}), and its fitting error is bounded below by the distances between neighbouring posterior means, minus the change the network can make between their embeddings; the family of marginals fixes these distances before any training.

We first verify the claim against its direct competitor, the direction of the levels from noisy to clean. Relabelling the embeddings leaves the objective unchanged, and a change in the fitting error can come only from the network. On a Gaussian mixture and CIFAR-10, reversing them flips the direction and keeps every neighbour while leaving the fitting error unchanged, whereas shuffling them disrupts the neighbourhoods and raises it. On the mixture and on CIFAR-10, the error broadly grows with the shuffle window. Embeddings learned with no imposed order place neighbouring levels near one another.

Second, we rule out sharing between distant levels, which the first test leaves in place, as the source of the benefit. Dividing the levels into contiguous ranges removes all sharing between the ranges and keeps the neighbours within them, and separate networks that keep the shared network's width fit as well as it. With the parameters held fixed, moving whole blocks of embeddings, which breaks adjacency at only as many places as such a division, costs less than 40\% of a full shuffle. We therefore show that the order of the noise levels enters a diffusion model through adjacency, whose cost the data fix before any training, with its direction being a design choice. 

\section{Objective and Noise-Level Embedding}
\label{sec:method}\label{sec:core}\label{sec:ownership}

Since the training objective contains no order of the noise levels, an order can enter a shared network only through its noise-level embedding and through what the noisy input reveals about its level. Let $x_0 \sim p_{\mathrm{data}}$ be a clean sample. For noise variances $a>b$, the Gaussian forward process draws a pair $y_b=x_0+\sqrt{b}\,\varepsilon$ and $y_a=y_b+\sqrt{a-b}\,\varepsilon'$, where $\varepsilon,\varepsilon'\sim\mathcal{N}(0,I)$ are independent of each other and of $x_0$. On a grid of noise variances $a_0>\dots>a_B$, write $y_k=y_{a_k}$ and $\hat x_k(y)=\mathbb{E}[x_0\mid y_{a_k}=y]$. The law of $y_{k+1}$ given $y_k$ depends on the data only through the posterior at level $k$ (Lemma~\ref{lem:one}), and drawing fresh noise gives the DDPM step \citep{ho2020ddpm}
\begin{equation}
p(y_{k+1} \mid y_k) = \mathcal{N}\!\big(r\,y_k + (1-r)\,\hat{x}_k(y_k),\;
\tau_k I\big), \qquad \tau_k = a_{k+1}(a_k - a_{k+1})/a_k,
\label{eq:stochlaw}
\end{equation}
with $r=a_{k+1}/a_k$. To fit $\hat x_k$, training draws a clean sample and noise at level $a_k$, and the transition likelihood reduces to a regression loss on the adjacent pair (Proposition~\ref{prop:rb}).

The full objective averages these local losses over noise levels. Since each term fits a prediction at one level, the unrestricted minimisers can be chosen independently. We define order invariance as the property that the objective is unchanged by any change of noise-level labels that preserves their training distributions and local losses (Figure~\ref{fig:invariance}). Write $\mathcal{L}_k$ for the expected loss at level $k$.

\begin{definition}[Sections and posterior means]
\label{def:sections}
For a function $f : \mathcal{X} \times K \to \mathcal{Y}$, the section $f(\cdot, k)$ is the prediction $f$ makes at noise level $k$. For a squared-error loss with regression target $t$, the posterior mean $F_k(y) = \mathbb{E}[\, t \mid y_k = y \,]$ is the minimiser of $\mathcal{L}_k$ over functions of the noisy input $y_k$.
\end{definition}

\begin{theorem}[Order invariance]
\label{thm:noorder}
Let $K$ be the set of noise levels, let $\pi$ be a distribution on $K$, and for each $k \in K$ let $\ell_k$ be a loss that
depends only on pairs drawn at noise level $k$. The population objective over
functions $f : \mathcal{X} \times K \to \mathcal{Y}$,
$\mathcal{L}(f) = \mathbb{E}_{k \sim \pi}\,
\mathbb{E}_{(x_0,\varepsilon)}[\, \ell_k(f(y_k, k);\, x_0,
\varepsilon) \,]$, is a function of the unordered family
$\{(k, \ell_k)\}_{k \in K}$. Its minimisers are the functions whose section at $\pi$-almost every noise level minimises $\mathcal{L}_k$, the same for every $\pi$ with the same null sets. On a finite $K$ every $\pi$ of full support has no null set, and the minimisers are the functions whose sections lie in $\arg\min \mathcal{L}_k$ at every $k$, the product set $\prod_{k \in K} \arg\min \mathcal{L}_k$.
\end{theorem}

Denoising score matching \citep{vincent2011connection}, $\varepsilon$-prediction \citep{ho2020ddpm}, $v$-prediction \citep{salimans2022progressive}, flow matching \citep{lipman2023flow} and the continuous-time objective \citep{song2021sde} all satisfy this hypothesis. The local loss $\ell_k$ retains the noise variance at level $k$.

\label{sec:axis}Theorem~\ref{thm:noorder} concerns functions assigned to individual noise levels. In a shared network $f_\theta$, training on pairs drawn at level $k$ also changes the prediction at level $j$ through parameters shared by the two levels. By Theorem~\ref{thm:noorder} the objective contains no order of the noise levels, so an order can enter a trained model only through its parameters. Apart from the noisy input and its $\sigma$-dependent scaling, the network distinguishes noise levels through a noise-level embedding $c_k$, and we write $f(x, c)$ for its prediction at input $x$ and embedding $c$, so that $f(\cdot, c_k)$ is its section at level $k$. In the models below, $c_k$ holds sinusoidal features of $\log\sigma_k$ at a fixed set of frequencies, so adjacent levels receive nearby embeddings. The embedding determines which noise levels are neighbours in the network (Figure~\ref{fig:orderenters}b). We test this by changing the embedding assigned to each noise level while keeping the training pairs and their true noise levels fixed. The objective therefore stays the same (Corollary~\ref{cor:relabel}).

\begin{corollary}[Relabelling]
\label{cor:relabel}
Let $s$ be a bijection of the noise levels, and let a model read noise level $k$
through the noise-level embedding $c_{s(k)}$ in place of $c_k$. As a function of what the
model assigns to each noise level, the objective of Theorem~\ref{thm:noorder}
and its set of minimisers are unchanged. A model class that can assign any
function to any embedding reaches the same minimum under every $s$.
\end{corollary}

We test these claims on an equally weighted mixture of 64 Gaussians in 64 dimensions, and on CIFAR-10 and ImageNet-256. The mixture's covariance spectrum is based on CIFAR-10 (Appendix~\ref{app:mog}). Its posterior means are available in closed form, so we measure excess error as the squared distance between the fitted function and the known posterior mean. For the mixture of Gaussian experiments, each network is a multilayer perceptron trained on fresh forward-process pairs. For images, we measure the FID of the composed models.

\section{Sharing Weights Across Noise Levels}
\label{sec:experiments}

By Theorem~\ref{thm:noorder}, separate networks with the same parameters would match one shared network if the noise levels were independent problems. We measure the gap between them, and how it changes with the width and the number of the networks, to find where the benefit of sharing comes from. A division removes every parameter the ranges share and keeps the neighbouring levels inside each range together. An interleaved division, used as a control, gives each of the $K$ networks every $K$-th transition, so that each network spans the whole range of noise levels. The training objective then becomes a sum of one loss per network, with each network minimising its own loss (Lemma~\ref{prop:own}). Because each transition ends at the marginal where the next one begins, the transition laws still compose into one sampler (Lemma~\ref{lem:pairing}). If the benefit came from the sharing that a division removes, the gap would grow with the number of networks and sit at their boundaries. If the useful sharing is between neighbouring levels, which a division keeps together, separate networks as wide as the shared one would fit as well as it, and the gap would follow the width each network gives up. Separate networks and their single-network baseline receive the same number of training steps.

\begin{figure}[t]
\centering
\includegraphics[width=\textwidth]{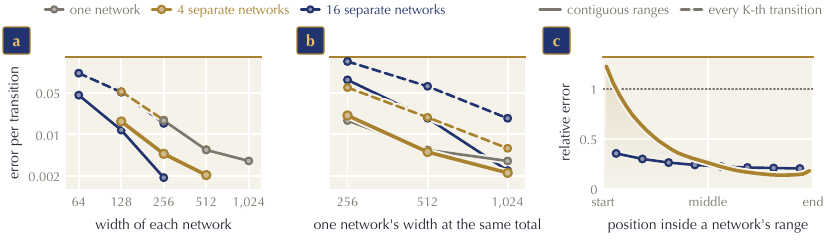}
\caption{Fitting with separate networks on the mixture experiments. Error is squared distance to the exact
posterior mean in the preconditioned space, averaged over transitions.
(a) Error versus the width of each network. (b) Error at the total parameter count of one network
of the width on the horizontal axis. Solid lines: contiguous ranges; dashed
lines: every $K$-th transition. (c) Error of 4 or 16 separate networks
divided by one network's error, all at width 256, calculated per transition and
averaged over ranges by position within the range. Means over 3 seeds.}
\label{fig:mogpar}
\end{figure}

\label{sec:fixedwidth}Separate networks as wide as the shared one match or improve on it. On the mixture, both 4 and 16 separate networks achieve lower error than one network of the same width, because each network covers a narrower range of noise levels (Figure~\ref{fig:mogpar}a). On CIFAR-10, the composed model matches the single-network baseline at every tested width, and the result also holds on ImageNet-256 with separate SiT-S/2 networks (Table~\ref{tab:splitting}). Where the image models differ, the composed model is better.

\begin{table}[t]
\centering\small
\caption{FID \citep{heusel2017gans} on 50k samples for $K$ separate networks on contiguous ranges, or every 4th transition in the
marked column, with equal training weight. Rows fix the total parameter count; cells give each network's width
and the actual count if it exceeds the target by over 2\%.
Dashes denote untrained models. CIFAR-10 \citep{krizhevsky2009learning} uses one evaluator; ImageNet-256 \citep{deng2009imagenet} uses SiT \citep{ma2024sit}
and the ADM evaluator \citep{dhariwal2021diffusion}.}
\label{tab:splitting}\label{tab:stripe}
\begin{tabular}{lcccc}
\toprule
 & \multicolumn{4}{c}{FID} \\
\cmidrule(lr){2-5}
Total & $K=1$ & $K=4$ & $K=4$, every 4th & $K=16$ \\
\midrule
0.875M & 19.82 (w16) & 30.71 (w8, 0.89M) & -- & 57.30 (w4, 0.92M) \\
3.5M & 9.45 (w32) & 14.53 (w16) & -- & 30.91 (w8, 3.57M) \\
14M  & 4.68 (w64) & 7.11 (w32) & 9.59 (w32) & 15.08 (w16) \\
56M  & 3.64 (w128) & 4.65 (w64) & 4.97 (w64) & 7.81 (w32) \\
\midrule
\multicolumn{5}{l}{\emph{ImageNet-256}} \\
33M & 66.39 (S/2) & 83.56 (XS/2) & -- & -- \\
132M & 39.92 (B/2) & 50.00 (S/2) & -- & 80.88 (XS/2) \\
\bottomrule
\end{tabular}
\end{table}

\begin{figure}[t]
\centering
\includegraphics[width=\textwidth]{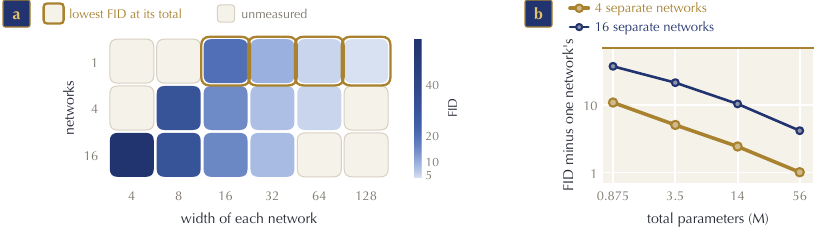}
\caption{(a) FID on a logarithmic colour scale. Columns increase the width of each network;
rows increase the number of separate networks. One column right or one row down multiplies total
parameters by 4. Empty cells are unmeasured, and a gold rim marks the lowest FID at each total. (b) FID of $K$ separate networks minus that of one
network at matched total parameters, plotted against that count on log axes.}
\label{fig:sweep}
\end{figure}

\label{sec:charge}At a fixed total parameter count, each separate network is narrower. On the mixture, dividing the transitions into contiguous ranges increases error for small networks and reduces it for large ones, for 4 and for 16 networks (Figure~\ref{fig:mogpar}b). On CIFAR-10, the FID cost of dividing the transitions at a fixed total parameter count comes from the smaller width of each network. The gap between $K$ separate networks and one network roughly halves for every 4-fold increase in the total parameter count (Figure~\ref{fig:sweep}b), and the FID of the separate networks follows the width of each network and depends little on whether they number 4 or 16 (Table~\ref{tab:splitting}).

\label{sec:boundaries}The gap does not come from the boundaries between networks. Beside a boundary, the loss moves by about 1\% in either direction, the interior of each range fits better than one network of the same width, and the boundary effect does not grow with the number of networks (Table~\ref{tab:seams}).

\begin{table}[t]
\centering\small
\caption{CIFAR-10, held-out denoising loss for $K$ separate networks of one width and one network of that width, after 60k training steps. We evaluate the loss with its training weight on 10k held-out pairs per transition with fixed noise, averaging separately over transitions beside a boundary between networks and the remaining transitions. The last column gives the percentage change from the single-network loss to the $K$-network loss.}
\label{tab:seams}
\begin{tabular}{llcccc}
\toprule
 & & & \multicolumn{3}{c}{held-out denoising loss} \\
\cmidrule(lr){4-6}
Networks & transitions & count & 1 network & $K$ separate networks & change \\
\midrule
4 of width 64 & beside a boundary & 6 & 0.1323 & 0.1334 & $+0.8\%$ \\
 & the rest & 122 & 0.4724 & 0.4593 & $-2.8\%$ \\
16 of width 32 & beside a boundary & 30 & 0.1603 & 0.1583 & $-1.3\%$ \\
 & the rest & 98 & 0.5568 & 0.5313 & $-4.6\%$ \\
\bottomrule
\end{tabular}
\end{table}

These results suggest that the gap comes from width, and a shared network can use its full width at every level because the targets of the levels are related. Every noisy marginal is the data density smoothed by a Gaussian, and smoothing the cleaner of two marginals by the difference in noise variance gives the noisier one, so all the posterior means $\{\hat x_k\}$ are determined by one data distribution. This relation holds between any two levels, but a finite network can use it only between levels whose embeddings are adjacent and whose posterior means lie close together, where a feature that helps fit one level also helps fit its neighbours. A separate network uses its features only on its own range and has fewer of them at a fixed total parameter count.

\section{Adjacency in the Noise-Level Embedding}
\label{sec:order}

A shared network organises this sharing through its noise-level embedding. The embedding fixes both the adjacency of the levels and their direction. A lower bound on the fitting error predicts that only adjacency matters, and reversed, shuffled and learned embeddings test the prediction.

\subsection{The embedding conflict bound}
\label{sec:bound}

We first bound the fitting error of a finite network from below, for any assignment of noise levels to embeddings. The bound is positive when the posterior means at adjacent embeddings differ more than the network's predictions can.

\begin{lemma}[Embedding conflict]
\label{lem:conflict}
Let the noise-level embeddings $c_1, \dots, c_B$ lie on a line, consecutive embeddings $\delta$
apart, and let noise level $s(i)$ be read through $c_i$. Let $F_k$ denote the
posterior mean of noise level $k$ in the model's output space, and let $f$ be
$\ell$-Lipschitz in its embedding in $L^2(\mu)$ for an input distribution
$\mu$. With $d_i = \lVert F_{s(i+1)} - F_{s(i)} \rVert_{L^2(\mu)}$ and
$e_k = \lVert f(\cdot, c_{s^{-1}(k)}) - F_k \rVert_{L^2(\mu)}$,
\[
\sum_{k} e_k^2 \;\ge\; \frac{1}{4} \sum_{i=1}^{B-1} \big(d_i - \ell\,\delta\big)_+^2 .
\]
\end{lemma}

When embeddings follow the noise-level order, $d_i$ is the distance between the posterior means of neighbouring levels, and refining the noise grid shrinks it together with the embedding spacing $\delta$. If this distance is no greater than the Lipschitz bound on the network's change between adjacent embeddings, the lower bound is zero. Shuffling can instead assign distant noise levels to adjacent embeddings, so the posterior means stay apart as a finer grid moves the embeddings closer together. Once $\ell\delta$ falls below these $d_i$, this separation gives a positive lower bound on error.

\subsection{Reversal and shuffle}
\label{sec:within}

Reversal and shuffle separate adjacency from direction. Reversing the embeddings changes their direction and keeps neighbouring noise levels together, whereas shuffling changes both. If direction mattered to fitting, reversal should raise the error as a shuffle does. The lower bound in Lemma~\ref{lem:conflict} is a sum over the pairs of neighbouring embeddings and depends on the assignment only through their distances $d_i$. Reversal keeps every pair and only reverses their order, which leaves the sum unchanged. With sinusoidal features, reversing an affine code is a fixed orthogonal map of the features that the first linear layer can absorb, so a reversed network can represent every function an ordered one can. We test whether fitted errors follow the same pattern and grow with the shuffle window, the number of consecutive levels among which the embeddings are shuffled. We also test whether the shuffle penalty persists as the network widens. An embedding learned for each noise level imposes no order and should match the ordered one if training places the embeddings of levels with similar posterior means near one another. Without an embedding, the network must read the noise level from the noisy state \citep{sun2025noise}.

In the Gaussian mixture experiments, the reversed embeddings fit the noise levels about as well as the ordered ones, and a full shuffle raises the error at every level (Figure~\ref{fig:orderenters}c). At noise levels between the trained ones, which no network saw, the error with ordered and reversed embeddings equals that at the trained levels, and with a shuffled embedding it is about 50 times larger or more (Table~\ref{tab:unseen}).

On CIFAR-10, we train one network of width 32 under each assignment of Table~\ref{tab:cifarorder}. FID varies as much between two trainings of one embedding as between any two embeddings (Table~\ref{tab:order}), so we compare the embeddings on the held-out denoising loss (Table~\ref{tab:cifarorder}). Reversing the embedding leaves the mean loss almost unchanged, and the learned and the continuous embeddings nearly match the ordered one. Shuffling raises the loss by a few per cent, far above the difference between two trainings. The distances $d_i$ of Lemma~\ref{lem:conflict} grow under the shuffle and keep their values under reversal, and the measured loss follows both. The rise is the same at widths 64 and 128. Most of it sits at the noisy end, where the noisy input carries least information about its noise level (Appendix~\ref{app:perlevel}). On a transformer, reversal changes the loss by less than twice the difference between two trainings, and the shuffle raises it by 3.5\%.

\begin{table}[t]
\centering\small\setlength{\tabcolsep}{4pt}
\caption{CIFAR-10, held-out denoising loss averaged over the 128 transitions, 10k pairs per transition with fixed noise, for one network trained for 60k steps. Left: width 32 under each noise-level embedding (continuous in $\sigma$, or one value per transition in noise-level order, reversed, learned, shuffled within windows of 2, 8 or 32 transitions, or fully shuffled); a second training with another seed in parentheses. The variation $\sum_i d_i^2$ from the training set's exact class-conditional posterior means is 1{,}187 for the ordered embedding, 3{,}296, 14{,}286 and 156{,}669 for windows of 2, 8 and 32, and 345{,}087 for the full shuffle. Right: ordered and shuffled embeddings at 3 widths, and the shuffle's rise. The last row replaces the U-Net with the transformer of Table~\ref{tab:hetero} (SiT, depth 12, width 256); its reversed embedding gives 0.4634.}
\label{tab:cifarorder}\label{tab:orderwidth}
\begin{tabular}[t]{lclc}
\toprule
\multicolumn{4}{c}{held-out denoising loss} \\
\cmidrule(lr){1-4}
Embedding & width 32 & Embedding & width 32 \\
\midrule
continuous & 0.4643 & window 2 & 0.4641 \\
ordered & 0.4640 (0.4642) & window 8 & 0.4673 \\
reversed & 0.4633 (0.4638) & window 32 & 0.4732 \\
learned & 0.4654 & shuffled & 0.4761 \\
\bottomrule
\end{tabular}\hspace{2em}
\begin{tabular}[t]{lccc}
\toprule
\multicolumn{4}{c}{held-out denoising loss} \\
\cmidrule(lr){1-4}
Width & ordered & shuffled & rise \\
\midrule
32 & 0.4640 & 0.4761 & 2.6\% \\
64 & 0.4569 & 0.4685 & 2.5\% \\
128 & 0.4493 (0.4491) & 0.4607 & 2.5\% \\
SiT & 0.4598 (0.4619) & 0.4759 & 3.5\% \\
\bottomrule
\end{tabular}
\end{table}

\subsection{Shuffle windows}
\label{sec:windows}

\begin{figure}[h]
\centering
\includegraphics[width=\textwidth]{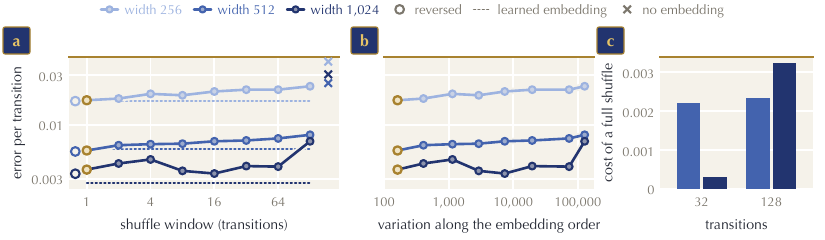}
\caption{Noise-level embedding comparison on the mixture experiments: one network of the listed width fits 128
transitions. (a) Mean excess error per transition versus shuffle-window size
(1: ordered; 128: full shuffle). Open circles: reversed order; dotted lines: learned
embedding per transition; crosses: no embedding. (b) Error versus variation of
the posterior means along the order of the embeddings, $\sum_i d_i^2$.
(c) Error increase from full shuffling over ordered embeddings on 32
and 128 transitions. Means over 5 seeds.}
\label{fig:order}
\end{figure}

A full shuffle breaks every neighbour at once. Shuffling within windows of increasing size breaks adjacency by degrees, and the bound predicts that the error grows with the variation of the posterior means along the embeddings. At widths 256 and 512 the error broadly grows with the shuffle window and follows this variation, $\sum_i d_i^2$, which we compute in closed form (Figure~\ref{fig:order}b). A full shuffle raises the error by about 40\% at widths 256 and 512. At width 1024 the rise varies by a factor of several across the 5 seeds, the partial shuffles scatter on both sides of the ordered embedding, and the reversed one fits a little better than the ordered one. On a grid of 32 transitions the embeddings lie 4 times further apart, so $\ell\delta$ covers more of each distance, and a shuffle adds less error, slightly less at width 512 and far less at width 1024 (Figure~\ref{fig:order}c).

On CIFAR-10 the held-out loss of a network of width 32 also grows with the shuffle window, in the order of the variation $\sum_i d_i^2$ from the training set's exact posterior means (Table~\ref{tab:cifarorder}). A window of 2 leaves the loss unchanged, and a window of 32 raises it by most of the full shuffle's rise.

\subsection{Learned embeddings}
\label{sec:learned}

The embeddings so far fix the order of the levels by hand. A learned embedding leaves the order to training and tests whether training finds adjacency on its own. The learned embedding nearly matches the ordered one at widths 256 and 512 and on CIFAR-10 (Table~\ref{tab:cifarorder}), and has lower error at width 1024. A network with no embedding has several times the error.

\begin{figure}[t]
\centering
\includegraphics[width=\textwidth]{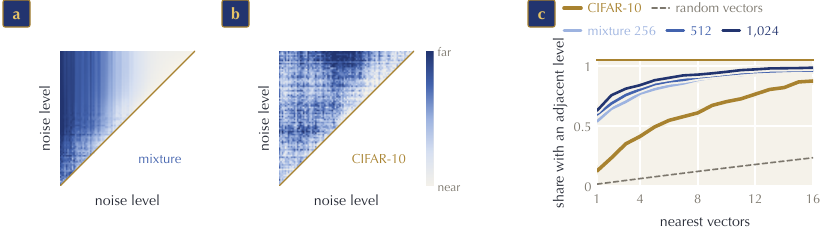}
\caption{Learned noise-level embeddings. (a) Distances between the learned vectors of the 128 transitions of one mixture network of width 512, from the noisiest level at the bottom left. (b) The same for one CIFAR-10 network of width 32. Each panel is on its own scale, clipped at the 2nd and 98th percentiles of its distances (light: near; dark: far). (c) The share of levels whose nearest learned vectors include an adjacent level, against the number of nearest vectors: the mixture at widths 256, 512 and 1024 (lighter to darker, means over 5 seeds), CIFAR-10, and random vectors.}
\label{fig:learned}
\end{figure}

On the mixture, neighbouring noise levels tend to have nearby learned vectors. At every width, most levels have an adjacent level among their nearest vectors, far more often than with random vectors (Figure~\ref{fig:learned}). On CIFAR-10, the learned vectors preserve neighbourhoods of noise levels but leave their order within each neighbourhood loose.

\section{Dividing the Noise Levels Among Networks}
\label{sec:division}

Within one network, the fitting error depends on which levels are neighbours in the embedding. Across networks, separate networks on contiguous ranges fit as well as one network of the same width, and at a fixed total parameter count their cost follows the width each network gives up (Section~\ref{sec:experiments}). A division of the noise levels among separate networks decides which levels share parameters at all, and adjacency predicts which divisions fit well. Networks that hold contiguous ranges should fit better than networks that hold interleaved levels. Breaking adjacency only at the $K-1$ boundaries of a division into $K$ ranges removes $K-1$ terms from the bound of Lemma~\ref{lem:conflict} and should cost far less than a shuffle, which makes every term large. The boundaries themselves can be set from the distances between neighbouring posterior means before training.

At a fixed total parameter count on the mixture, separate networks that each hold every $K$-th transition have several times the error of networks on contiguous ranges (Table~\ref{tab:embedding}), although at a fixed width they match or improve on a single network (Figure~\ref{fig:mogpar}a). A network that holds neighbouring levels fits a narrow part of the family of marginals, and an interleaved one must span all of it.

\begin{table}[t]
\centering\small
\caption{Excess error per transition on the mixture experiments, $\times 100$, mean $\pm$ standard deviation over 3 seeds. Rows fix the total parameter count by one network's width, which the $K$ separate networks share (Appendix~\ref{app:mog}); they hold contiguous ranges unless marked every 4th or 16th transition. The noise-level embedding is sinusoidal, learned (16 trained parameters per transition) or a fixed random vector.}
\label{tab:embedding}\label{tab:mogpar}
\begin{tabular}{llccccc}
\toprule
 & & \multicolumn{5}{c}{excess error $\times 100$, $K$ networks} \\
\cmidrule(lr){3-7}
Width of 1 network & embedding & 1 & 4 & 4, every 4th & 16 & 16, every 16th \\
\midrule
256 & sinusoidal & 1.71\,{\scriptsize$\pm$0.06} & 2.08\,{\scriptsize$\pm$0.03} & 6.12\,{\scriptsize$\pm$0.03} & 8.24\,{\scriptsize$\pm$0.02} & 16.83\,{\scriptsize$\pm$0.08} \\
 & learned & 1.71\,{\scriptsize$\pm$0.05} & 2.07\,{\scriptsize$\pm$0.08} & 5.54\,{\scriptsize$\pm$0.05} & 7.94\,{\scriptsize$\pm$0.03} & 16.68\,{\scriptsize$\pm$0.10} \\
 & random & 2.07\,{\scriptsize$\pm$0.00} & 2.23\,{\scriptsize$\pm$0.12} & 6.84\,{\scriptsize$\pm$0.66} & 7.94\,{\scriptsize$\pm$0.06} & 16.73\,{\scriptsize$\pm$0.12} \\
\midrule
512 & sinusoidal & 0.54\,{\scriptsize$\pm$0.01} & 0.50\,{\scriptsize$\pm$0.01} & 1.93\,{\scriptsize$\pm$0.04} & 1.87\,{\scriptsize$\pm$0.02} & 6.43\,{\scriptsize$\pm$0.03} \\
 & learned & 0.55\,{\scriptsize$\pm$0.01} & 0.50\,{\scriptsize$\pm$0.00} & 1.85\,{\scriptsize$\pm$0.06} & 1.78\,{\scriptsize$\pm$0.02} & 5.85\,{\scriptsize$\pm$0.02} \\
 & random & 0.72\,{\scriptsize$\pm$0.05} & 0.57\,{\scriptsize$\pm$0.02} & 2.35\,{\scriptsize$\pm$0.31} & 1.87\,{\scriptsize$\pm$0.01} & 6.58\,{\scriptsize$\pm$0.02} \\
\bottomrule
\end{tabular}
\end{table}

\begin{figure}[t]
\centering
\includegraphics[width=\textwidth]{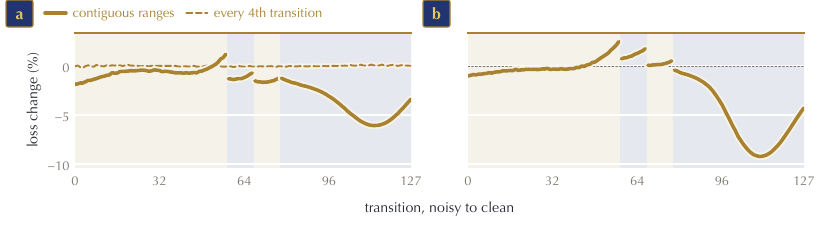}
\caption{CIFAR-10, loss change for 4 separate networks relative to one network of the same width,
from noisiest to cleanest transition: separate networks' loss minus single-network loss,
divided by single-network loss, in percent. Separate networks hold contiguous ranges;
shading marks the ranges of the second and fourth networks. (a) 14M, width 32; the dashed line shows 4 separate networks
holding every 4th transition. (b) 56M, width 64. Loss uses training weights
and 10k held-out pairs per transition with fixed noise.}
\label{fig:profile}
\end{figure}

CIFAR-10 shows the same cost of interleaving. A network that holds every 4th transition shares its parameters among levels that are not neighbours and must span the full range of noise levels, as a single network does. Its denoising loss is within 0.5\% of that of one network of the same width at every transition (Figure~\ref{fig:profile}a), and the composed model's FID is higher than with contiguous ranges, by a little at 56M and by more at 14M. At the same width, its FID is close to that of a single network (Table~\ref{tab:splitting}, the column of every 4th transition).

For each data set, the highest relative loss in every range occurs at the end closest to the same transition. On the mixture experiments, this place is the peak of the single network's error, near transition 35 (Figure~\ref{fig:orderenters}c). Figure~\ref{fig:mogpar}c averages the ratio to one network over the ranges, by position within a range, and the ratio is highest at the noisy end, since 3 of the 4 ranges lie on the clean side of that peak. The start of the third range is the only boundary at which the separate networks have more error than one network, several times as much. On CIFAR-10, the place is the start of the last range: loss increases toward the clean end of the first 3 ranges and peaks at the noisy end of the last (Figure~\ref{fig:profile}).

The placement of the boundaries therefore matters, and the family of marginals can set it before training. For each transition, we measure how far the exact posterior mean moves between its two noise levels (Appendix~\ref{app:protocols}), and we divide the grid into 4 contiguous ranges that each hold 25\% of the summed movement, keeping every network's width and training. On the mixture, these boundaries lower the error by 40\% to 50\% at the two larger parameter counts and raise it slightly at the smallest (Table~\ref{tab:cuts}). On CIFAR-10 at 14M, the boundaries from the training set's exact posterior mean lower the held-out loss by a few per cent, almost all of it in the cleanest range. Applying each data set's boundaries to the other gives asymmetric results. On the mixture, CIFAR-10's boundaries also lower the error at the two larger parameter counts, but less than the mixture's own. On CIFAR-10, the mixture's boundaries leave the held-out loss almost unchanged.

\begin{table}[t]
\centering\small
\caption{Four separate networks on contiguous ranges, with the widths and training of the four-network models of Figure~\ref{fig:mogpar}b and Table~\ref{tab:splitting}; only the boundaries move. The equal split gives the mixture's networks 32 transitions each and CIFAR-10's equal shares of EDM's training distribution (58, 10, 10 and 50 transitions). The other rows place the boundaries at 25\%, 50\% and 75\% of the summed movement of the exact posterior mean, computed on the same or on the other data set. Mixture: mean excess error per transition $\pm$ standard deviation, $\times 100$, over 3 seeds, at the total parameter count of one network of width 256, 512 or 1024. CIFAR-10, 14M: held-out denoising loss averaged over the 128 transitions, 10k pairs per transition with fixed noise.}
\label{tab:cuts}
\begin{tabular}{lcccc}
\toprule
 & \multicolumn{3}{c}{Mixture, excess error $\times 100$} & CIFAR-10, 14M \\
\cmidrule(lr){2-4}\cmidrule(lr){5-5}
Boundaries & 256 & 512 & 1024 & held-out loss \\
\midrule
equal split & 2.075\,{\scriptsize$\pm$0.030} & 0.503\,{\scriptsize$\pm$0.007} & 0.224\,{\scriptsize$\pm$0.006} & 0.4530 \\
own posterior mean & 2.114\,{\scriptsize$\pm$0.026} & 0.298\,{\scriptsize$\pm$0.004} & 0.117\,{\scriptsize$\pm$0.002} & 0.4414 \\
other data set's posterior mean & 2.398\,{\scriptsize$\pm$0.048} & 0.333\,{\scriptsize$\pm$0.004} & 0.124\,{\scriptsize$\pm$0.000} & 0.4539 \\
\bottomrule
\end{tabular}
\end{table}

Reordering whole blocks within one network changes only the pairs of neighbouring levels at the boundaries between blocks, which are the pairs a division into as many ranges separates (Table~\ref{tab:blocks}). At widths 256 and 512, 4 moved blocks raise the error by about 10\% of the full shuffle's rise and 16 by about 33\%. Per transition, the rise where neighbours changed is several times the rise elsewhere. A division, which separates the same pairs, lowers the error at a fixed width (Section~\ref{sec:experiments}).

\begin{table}[t]
\centering\small\setlength{\tabcolsep}{4.5pt}
\caption{Mixture, one network over 128 transitions: mean excess error per transition $\pm$ standard deviation, $\times 100$, over 5 seeds, with the embeddings in order, reversed, with 4 or 16 contiguous blocks moved and each block's order kept (3 or 15 pairs of neighbours separated), or fully shuffled (127 pairs). The ratios divide the rise per transition where neighbours changed by the rise elsewhere; a dash marks a negative rise elsewhere.}
\label{tab:blocks}
\begin{tabular}{lccccccc}
\toprule
 & \multicolumn{5}{c}{excess error $\times 100$} & \multicolumn{2}{c}{ratio} \\
\cmidrule(lr){2-6}\cmidrule(lr){7-8}
Width & ordered & reversed & 4 blocks & 16 blocks & full shuffle & 4 blocks & 16 blocks \\
\midrule
256 & 1.72\,{\scriptsize$\pm$0.10} & 1.69\,{\scriptsize$\pm$0.08} & 1.77\,{\scriptsize$\pm$0.07} & 1.96\,{\scriptsize$\pm$0.07} & 2.35\,{\scriptsize$\pm$0.29} & 9.5 & 2.8 \\
512 & 0.56\,{\scriptsize$\pm$0.01} & 0.55\,{\scriptsize$\pm$0.02} & 0.59\,{\scriptsize$\pm$0.01} & 0.64\,{\scriptsize$\pm$0.02} & 0.80\,{\scriptsize$\pm$0.04} & 9.5 & 3.0 \\
1024 & 0.37\,{\scriptsize$\pm$0.02} & 0.34\,{\scriptsize$\pm$0.01} & 0.43\,{\scriptsize$\pm$0.07} & 0.34\,{\scriptsize$\pm$0.08} & 0.69\,{\scriptsize$\pm$0.24} & 6.8 & -- \\
\bottomrule
\end{tabular}
\end{table}

Dividing transitions among separate networks should reduce the penalty for unordered embeddings, since each network shares parameters across fewer levels. On the mixture networks, we replace sinusoidal embeddings with learned or fixed random vectors (Table~\ref{tab:embedding}). Learned vectors match sinusoidal embeddings on one network and fit up to 10\% better once the levels are divided. With one network, random vectors raise error by about 20\% at width 256 and 33\% at width 512; with 16 networks they lie within a few per cent of sinusoidal embeddings.

Once the ranges share no parameters, the transition law is the only interface between the networks, and the output distribution changes by at most the summed change of the laws a replaced network supplies (Theorem~\ref{prop:tes}). On CIFAR-10, retraining one of the 16 separate networks with a fresh seed changes the composed sampler's FID by only a few hundredths, and a transformer and 3 U-Nets trained as 4 separate networks on different machines compose to within 1 FID point of 4 separate networks with the same architecture (Table~\ref{tab:modular}).

\begin{table}[t]
\centering\small
\caption{CIFAR-10, FID over 50k samples on one evaluator, for composed models whose networks are retrained or replaced. The 4 architectures are a transformer and 3 U-Nets, trained on different machines, in place of 4 U-Nets of width 64.}
\label{tab:modular}
\begin{tabular}{lcccc}
\toprule
 & \multicolumn{2}{c}{16 networks of width 32} & \multicolumn{2}{c}{4 networks} \\
\cmidrule(lr){2-3}\cmidrule(lr){4-5}
 & as trained & one retrained, fresh seed & width 64 & 4 architectures \\
\midrule
FID & 7.81 & 7.79 & 4.65 & 5.43 \\
\bottomrule
\end{tabular}
\end{table}

\FloatBarrier
\section{Related Work}
\label{sec:related}

One line of work reads the noise levels as separate tasks and keeps the shared network. \citet{hang2023minsnr} describe diffusion training as multi-task learning whose tasks conflict, and reweight the loss by the signal-to-noise ratio. \citet{go2023negative} measure negative transfer between noise levels, find that the affinity between two levels is highest for neighbours and falls with their distance, and group the levels into intervals; \citet{park2024dtr} give each noise level its own subset of channels and share more channels between adjacent levels. We study a different object, which levels are neighbours in the noise-level embedding. With the tasks fixed, we show that the benefit of sharing rests on this adjacency in the embedding while the direction of the levels is a design choice, bound the fitting error by it, and test it by reversing, shuffling and dividing the levels. \citet{sun2025noise} remove the noise-level input and let the network infer the noise level from the noisy state, and the one network still covers every noise level. Unlike these methods, we compare one shared network with separately trained networks at a matched width and at a matched total parameter count, and assign each network either neighbouring or interleaved levels. These controls separate the effect of network width from that of level grouping (Sections~\ref{sec:experiments} and~\ref{sec:division}).

Further related work is discussed in Appendix~\ref{app:related}.

\section{Conclusion}
\label{sec:limitations}

The training objective of a diffusion model is pointwise in the noise level, yet on images one network shared across the levels outperforms separate networks with the same total number of parameters. The benefit of sharing rests on adjacency in the noise-level embedding, since a lower bound on the fitting error of a shared network grows with the distances between the posterior means of neighbouring levels. Reversing the embeddings keeps these distances and the fitting error, whereas shuffling them raises both. Separate networks that keep neighbouring levels together fit as well as one network of their width, and within one network, breaking adjacency only at the boundaries of such a division costs less than 40\% of a full shuffle. The bound predicts which embeddings fit worse but not by how much; our image experiments use small backbones without guidance; and we test only first-order samplers. The evidence for direction being a design choice is limited to the settings we test. The order of the noise levels therefore enters a diffusion model through adjacency, with its direction being a design choice.

\setlength{\belowcaptionskip}{2pt}

\section*{Ethics Statement}
We train on CIFAR-10 and ImageNet, both public. The models are image generators at research scale. Our results concern how the training of such models is organised and do not extend what these models can produce.

\section*{Reproducibility Statement}
Appendix~\ref{app:laws} gives assumptions and proofs; Appendix~\ref{app:mog} specifies the mixture data, networks and sampler; Appendix~\ref{app:protocols} gives the CIFAR-10 and ImageNet training and evaluation protocols. Supplementary code trains and composes the networks and reproduces the figures and tables.

\section*{AI Use Statement}
We used generative AI tools to polish the manuscript and check novelty against existing publications. We also used them for coding and for checking derivations. We have reviewed all AI-assisted work and take responsibility for the content of the paper.

\bibliography{btdm}
\bibliographystyle{preprint}

\appendix

\section{Statements and Proofs}
\label{app:laws}

\begin{lemma}[The law between two noise levels]
\label{lem:one}
Let $a > b$ be two noise levels, $r = b/a$ and $\tau = r(1-r)\,a$, and condition
on $y_a = y$. Given $x_0$ the state $y_b$ is Gaussian with mean
$r\,y + (1-r)\,x_0$ and covariance $\tau I$. The law of $y_b$ is therefore
$\int \mathcal{N}\big(r\,y + (1-r)\,x_0,\ \tau I\big)\,p(x_0 \mid y_a = y)\,
\mathrm{d}x_0$, with mean $r\,y + (1-r)\,\mathbb{E}[x_0 \mid y_a = y]$.
\end{lemma}
\begin{proof}
Under the forward process $y_b = x_0 + \sqrt{b}\,\varepsilon$ and
$y_a = y_b + \sqrt{a-b}\,\varepsilon'$, with $\varepsilon$ and $\varepsilon'$
standard normal and independent of $x_0$. Given $x_0$ the pair is jointly
Gaussian, and conditioning $y_b$ on $y_a = y$ gives the mean
$x_0 + (b/a)(y - x_0) = r\,y + (1-r)\,x_0$ and the covariance
$b(a-b)/a\, I = \tau I$. Integrating $x_0$ against $p(x_0 \mid y_a = y)$
gives the stated law, and its mean follows by linearity. The couplings
\eqref{eq:stochlaw} and \eqref{eq:diraclaw} replace $x_0$ by $\hat x_k(y)$,
drawing fresh noise for the first and reusing the drawn noise for the
second. These two sampling steps depend on the data only through $\hat x_k$.
\end{proof}

On the grid of Section~\ref{sec:method}, the DDPM step \eqref{eq:stochlaw} draws fresh noise. Reusing the noise that produced $y_k$ gives the DDIM step \citep{song2021ddim}
\begin{equation}
y_{k+1} = \hat{x}_k + \sqrt{r}\,\big(y_k - \hat{x}_k\big).
\label{eq:diraclaw}
\end{equation}

Both steps depend on the data only through $\hat x_k$, which Tweedie's formula relates to the score of the noisy marginal \citep{robbins1956empirical, efron2011tweedie}. The step \eqref{eq:stochlaw} replaces the posterior in Lemma~\ref{lem:one} by its mean: the two share a mean and differ in covariance by $(1-r)^2\,\mathrm{Cov}(x_0\mid y_a=y)$, which vanishes relative to $\tau$ as the transition shortens.

\begin{proposition}[One estimator]
\label{prop:rb}
For the transition law \eqref{eq:stochlaw} the expectation of the transition
negative log-likelihood over the forward-process noise equals
$\frac{1-r}{2r}\,\lVert x_0 - \hat{x}_k(y_k)\rVert^2 / a_k$ up to a
constant for the transition.
\end{proposition}
\begin{proof}
The negative log-likelihood of \eqref{eq:stochlaw} at a state
$y_{k+1}$ is $\lVert y_{k+1} - r\,y_k - (1-r)\,\hat x_k \rVert^2 /
(2\tau_k)$ plus a constant for the transition. Under the forward process the drawn
state is $y_{k+1} = r\,y_k + (1-r)\,x_0 + \sqrt{\tau_k}\,
\varepsilon$ with standard normal $\varepsilon$, so the quadratic is
$\lVert (1-r)(x_0 - \hat x_k) + \sqrt{\tau_k}\,\varepsilon
\rVert^2 / (2\tau_k)$. The cross term vanishes in expectation and the
$\varepsilon$ term contributes $d/2$, a constant for the transition. With
$\tau_k = r(1-r)\,a_k$ the coefficient $(1-r)^2/(2\tau_k)$ reduces
to $(1-r)/(2r a_k)$, the stated weight. Minimising the expectation in
$\hat x$ therefore gives weighted regression with the clean sample as the target.
\end{proof}

\begin{lemma}[Composition of joint laws]
\label{lem:pairing}
For noise levels $a > b$, let $\Gamma_{a,b}$ be a joint law of $(y_a, y_b)$ with
marginals $m_a$ and $m_b$, and let $\hat p(\cdot \mid y_a)$ be its
conditional. A sampler that arrives at noise level $a$ with law $m_a$ and applies
$\hat p$ reaches noise level $b$ with law $m_b$. Consider a sequence of joint
laws $\Gamma_{a_0,a_1}, \dots, \Gamma_{a_{B-1},a_B}$ whose adjacent marginals
agree. Starting from $y_0 \sim m_{a_0}$, a sampler that applies their conditionals
in sequence ends with law $m_{a_B}$, whatever process produced each $\Gamma$.
\end{lemma}
\begin{proof}
After the transition, the sampler has law $\int \hat p(\cdot \mid y_a)\,
m_a(\mathrm{d}y_a)$. Since $\hat p$ is the conditional of $\Gamma_{a,b}$
and $m_a$ is its first marginal, this integral equals the second marginal $m_b$.
At transition $k+1$, the sampler has the marginal of $\Gamma_{a_k,a_{k+1}}$
at $a_{k+1}$. By hypothesis, this is the first marginal of
$\Gamma_{a_{k+1},a_{k+2}}$, so the next conditional produces the required marginal
at the next noise level. Repeating this argument proves the claim at $a_B$. Nothing in the argument refers to how any
$\Gamma$ was produced.
\end{proof}

Since every forward-process $\Gamma$ has the marginals $p_{\mathrm{data}} \ast \mathcal{N}(0, a I)$, the forward process satisfies the condition. The lemma also allows pairs from any other family of joint laws whose adjacent marginals agree, under one condition: the sampler must apply the conditional of the joint law the pairs came from.

\noindent\textbf{Theorem~\ref{thm:noorder} (Order invariance).}
\emph{Let $K$ be the set of noise levels, let $\pi$ be a distribution on $K$, and for each $k \in K$ let $\ell_k$ be a loss that
depends only on pairs drawn at noise level $k$. The population objective over
functions $f : \mathcal{X} \times K \to \mathcal{Y}$,
$\mathcal{L}(f) = \mathbb{E}_{k \sim \pi}\,
\mathbb{E}_{(x_0,\varepsilon)}[\, \ell_k(f(y_k, k);\, x_0,
\varepsilon) \,]$, is a function of the unordered family
$\{(k, \ell_k)\}_{k \in K}$. Its minimisers are the functions whose section at $\pi$-almost every noise level minimises $\mathcal{L}_k$, the same for every $\pi$ with the same null sets. On a finite $K$ every $\pi$ of full support has no null set, and the minimisers are the product set $\prod_{k \in K} \arg\min \mathcal{L}_k$.}
\begin{proof}
Since $\ell_k$ depends only on noise level $k$, we can write
$\mathcal{L}(f) = \int_K \mathcal{L}_k(f(\cdot,k))\,
\mathrm{d}\pi(k)$. For a measurable bijection $\sigma$ of $K$ set $(\sigma f)(\cdot,k) =
f(\cdot, \sigma^{-1}k)$, $\ell'_k = \ell_{\sigma^{-1}k}$ and $\pi' =
\sigma_\ast \pi$. Then $\mathcal{L}'(\sigma f) = \int_K
\mathcal{L}_{\sigma^{-1}k}(f(\cdot, \sigma^{-1}k))\,
\mathrm{d}(\sigma_\ast\pi)(k) = \mathcal{L}(f)$, so $\mathcal{L}$ is a
function of the unordered family. An ordering, an adjacency, an
orientation or a metric on $K$ is changed by some relabelling that
leaves $\mathcal{L}$ unchanged, so $\mathcal{L}$ uses none of them. The noise
variance $a_k$ belongs to the term $\ell_k$ and moves with it under every
relabelling. For the minimisers,
$\mathcal{L}(f) \ge \int_K \inf_g \mathcal{L}_k(g)\, \mathrm{d}\pi(k)$
with equality if and only if $f(\cdot,k)$ attains the inner infimum
for $\pi$-almost every $k$. The values $f(\cdot,k)$ at distinct noise levels are unconstrained by one another, so the bound is attained by a minimiser at each noise level separately, and the minimising set consists of the functions that minimise $\mathcal{L}_k$ at $\pi$-almost every $k$. This set depends on $\pi$ only through its null sets. On a finite $K$ a distribution of full support has no null set, so the set is the product $\prod_k \arg\min \mathcal{L}_k$ and is the same for every such $\pi$.
\end{proof}

\begin{figure}[t]
\centering
\includegraphics[width=\textwidth]{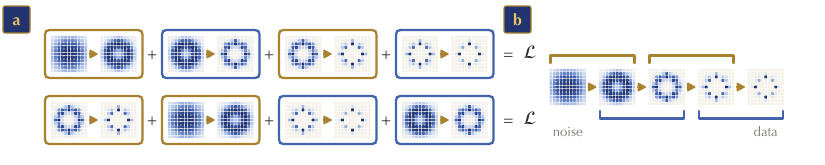}
\caption{What order invariance separates. Each domino is one transition law between the
marginals at two adjacent noise levels, fitted by its own network (frame colour), and each face is
the density of a mixture of 8 Gaussians smoothed at that level. (a) The objective adds one term
per law, computed from that law's pairs alone, and gives the same sum in noise order (top) and in a
shuffled order (bottom). (b) A sampler joins the laws where their marginals agree: each interior
marginal ends one law (above) and starts the next (below).}
\label{fig:invariance}
\end{figure}

\noindent\textbf{Where an order enters (Section~\ref{sec:axis}).}
\emph{Let training descend $\mathcal{L}(\theta) = \mathbb{E}_{k \sim
\pi}\, \mathcal{L}_k(f_\theta(\cdot, k)) + \mathcal{M}(\theta) +
R(\theta)$ for a parameterisation $f_\theta$, where $\mathcal{M}$
collects every loss term that depends on pairs from more than one noise level and
$R$ any regulariser. A training update using pairs drawn at noise level $k$ can change $f_\theta(\cdot,
j)$, $j \neq k$, only through a parameter that enters both
$f_\theta(\cdot, j)$ and $f_\theta(\cdot, k)$, through $\mathcal{M}$,
or through $R$. With the parameters disjoint
across noise levels, $\mathcal{M} = 0$ and $R$ separable over the noise levels,
the update due to these pairs leaves the predictions at the other noise level unchanged.}
\begin{proof}
A descent step of size $\eta$ on $\mathcal{L}(\theta)$ moves the
parameters by $\Delta\theta = -\eta\big(\int_K (\partial_\theta
f_\theta(\cdot,k))^{\!\top} \nabla_{\!f} \mathcal{L}_k\, \mathrm{d}\pi(k)
+ \nabla_\theta \mathcal{M} + \nabla_\theta R\big)$, and to first order
moves $f_\theta(x, j)$ by $\partial_\theta f_\theta(x,j)\, \Delta\theta$.
The noise level-$k$ term contributes $-\eta\, \partial_\theta f_\theta(x,j)\,
(\partial_\theta f_\theta(\cdot,k))^{\!\top} \nabla_{\!f}
\mathcal{L}_k$, and the operator $\partial_\theta f_\theta(x,j)\,
(\partial_\theta f_\theta(\cdot,k))^{\!\top} = \sum_i \partial_{\theta_i}
f_\theta(x,j) \otimes \partial_{\theta_i} f_\theta(\cdot,k)$ is a sum
over the coordinates $\theta_i$ that enter both noise levels. The pairs drawn at
noise level $k$ contribute to $\Delta\theta$ through this term and through
$\nabla_\theta \mathcal{M}$ when $\mathcal{M}$ depends on noise level $k$.
There are no other contributions from these pairs: $R$ is independent of the
training pairs, and every other $\mathcal{L}_{k'}$ depends on its own noise level.
When no coordinate enters two noise levels, the sum is empty. Setting
$\mathcal{M} = 0$ removes its term. If $R$ is separable, its gradient with respect
to the parameters of noise level $j$ depends on that noise level alone. The noise level-$j$ term remains. The same
computation holds along a gradient flow.
\end{proof}

\begin{definition}[Finite transition model]
\label{def:ftm}
Let $a_0 > \dots > a_B$ be a grid of noise variances and call the move from noise level $a_k$ to $a_{k+1}$ a
transition. A finite transition model carries one law $\hat p_k$ for each of
the $B$ transitions, each with parameters of its own. A sample is drawn from
$y_0 \sim \mathcal{N}(0, a_0 I)$ by applying the laws in the order of the
noise levels.
\end{definition}

\begin{lemma}[Disjoint parameters realise the factorisation]
\label{prop:own}
Let the transitions be grouped into sets $S_1, \dots, S_W$ and the parameters carry disjoint blocks $\theta = (\theta_1, \dots, \theta_W)$, one per group, with $\hat p_k$ reading $\theta_w$ only for $k \in S_w$. Let $\mathcal{L}(\theta) = \sum_w \mathcal{L}_w(\theta_w)$ with $\mathcal{L}_w = \sum_{k \in S_w} \mathcal{L}_k$ the group's terms of Proposition~\ref{prop:rb}. Then $\nabla_{\theta_w} \mathcal{L} = \nabla_{\theta_w} \mathcal{L}_w$ for every $w$, and a tuple minimises $\mathcal{L}$ if and only if each block minimises its own group's sum. With one transition per group the minimisers are the per-transition minimisers of Theorem~\ref{thm:noorder}.
\end{lemma}
\begin{proof}
By Proposition~\ref{prop:rb}, the transition-$k$ term is an expectation over
pairs drawn at noise level $a_k$ of a quantity depending on $x_0$, $y_{a_k}$
and $\hat x_k$. For $k \in S_w$, the estimator $\hat x_k$ depends only on $\theta_w$.
Thus $\nabla_{\theta_v} \mathcal{L}_k$ vanishes identically for $v \neq w$.
Differentiating the sum term by term gives the gradient identity. A sum of functions of disjoint arguments is minimised exactly when each summand is minimised in its own argument, which is the second claim. With one transition per group each summand is one $\mathcal{L}_k$, whose minimisers are those of Theorem~\ref{thm:noorder}.
\end{proof}

\noindent\textbf{Theorem~\ref{prop:tes} (Modularity).}
\emph{Let the transitions carry disjoint parameters and let the sampler draw
$y_{k+1} \sim \hat p_k(\cdot \mid y_k)$ for $k = 0, \dots, B-1$ from
$y_0 \sim \mathcal{N}(0, a_0 I)$. A second sampler from the same prior
that replaces $\hat p_j$ by $\hat p'_j$ for every $j$ in a set $J$ of
transitions satisfies
\[
\mathrm{TV}\big(\mathrm{law}(y_B),\ \mathrm{law}(y'_B)\big)
\;\le\; \sum_{j \in J} \mathbb{E}\,\mathrm{TV}\big(\hat p_j(\cdot \mid
y_j),\ \hat p'_j(\cdot \mid y_j)\big),
\]
where the expectation is taken under the marginal at transition $j$ of
the chain with all replacements before $j$ already applied.}
\begin{proof}
Order $J$ as $j_1 < \dots < j_m$. Let $H_i$ be the chain with
the transitions $j_1, \dots, j_i$ replaced. Then $H_0$ is the first
sampler and $H_m$ the second. Consecutive chains $H_{i-1}$ and $H_i$
differ only in the kernel at transition $j_i$. They start from the same prior
and apply the same kernels before this transition. The state $y_{j_i}$ therefore
has the same law $\mu$ under both chains. The law of $y_{j_i + 1}$
is $\mu \hat p_{j_i}$ under $H_{i-1}$ and $\mu \hat p'_{j_i}$ under
$H_i$. By convexity of total variation, $\mathrm{TV}(\mu \hat
p_{j_i}, \mu \hat p'_{j_i}) \le \mathbb{E}_{y \sim \mu}\,
\mathrm{TV}(\hat p_{j_i}(\cdot \mid y), \hat p'_{j_i}(\cdot \mid y))$.
The transitions after $j_i$ apply the same kernels to both chains. Applying a
common kernel cannot increase total variation, so the triangle inequality over
$i = 1, \dots, m$ gives the bound. Here $\mu$ is the marginal at transition $j_i$
of $H_{i-1}$, the chain in which all earlier replacements have been applied.
\end{proof}

\noindent\textbf{Grouping and retraining (Section~\ref{sec:axis}).}
\emph{In a finite transition model, every grouping of the transitions trained independently
has the same minimisers, and at those minimisers the laws, applied in the grid's order, compose
to the same model, whether a group holds a contiguous range or every $K$-th transition.
Retraining one transition changes only that transition law. The change in the
sample distribution is bounded by the expected total variation distance between
the original and retrained laws, as in Theorem~\ref{prop:tes}.}
\begin{proof}
For a grouping of transitions into blocks, each block is trained by gradient
descent on the sum of its own loss terms. By Lemma~\ref{prop:own} with one transition per block the minimisers of that sum are the per-transition minimisers, so every grouping has the
same minimising laws. The sampler applies the laws in the order of the noise levels and its
output is a function of the laws only, so the composed model at the minimisers is the
same for every grouping, whether or not the transitions of a block adjoin.
Retraining transition $j$ updates only $\theta_j$ to reduce
$\mathcal{L}_j(\theta_j)$, since the parameter blocks are disjoint. Every other transition law therefore remains unchanged. Theorem~\ref{prop:tes} with $J = \{j\}$ then bounds the change in
the output distribution.
\end{proof}

\noindent\textbf{Corollary~\ref{cor:tes} (Error of the composed model).}
\emph{With $J$ every transition, $\hat p'_j$ the transition laws $p_j$ of the
forward process and the prior $p_{a_0}$, Theorem~\ref{prop:tes}
gives $\mathrm{TV}(\mathrm{law}(y_B), p_{\mathrm{data}} \ast
\mathcal{N}(0, a_B I)) \le \mathrm{TV}(\mathcal{N}(0, a_0 I), p_{a_0})
+ \sum_k \mathbb{E}\,\mathrm{TV}(\hat p_k, p_k)$, whose sum vanishes with exact laws at every transition count and whose first term vanishes as $a_0$ grows.}
\begin{proof}
The chain of the true transition laws from $p_{a_0}$ composes by the
Chapman--Kolmogorov equation to $p_{\mathrm{data}} \ast \mathcal{N}(0,
a_B I)$. The two samplers differ in the prior as well as in every transition,
and the prior adds one term by the argument of Theorem~\ref{prop:tes}
at transition $0$, since a common kernel does not increase total variation.
With exact laws every term of the sum is zero.
\end{proof}

\noindent\textbf{Corollary~\ref{cor:relabel} (Relabelling).}
\emph{Let $s$ be a bijection of the noise levels, and let a model read noise level $k$
through the noise-level embedding $c_{s(k)}$ in place of $c_k$. As a function of what the
model assigns to each noise level, the objective of Theorem~\ref{thm:noorder}
and its set of minimisers are unchanged. A model class that can assign any
function to any embedding reaches the same minimum under every $s$.}
\begin{proof}
Write $h(\cdot, k) = f(\cdot, c_{s(k)})$ for the prediction function at noise
level $k$ after relabelling. The objective of Theorem~\ref{thm:noorder} is
$\mathbb{E}_{k \sim \pi}\,\mathcal{L}_k(h(\cdot, k))$. It depends on $h$ alone,
so its minimisers over $h$ form the same product set for every $s$.
Since $s$ is a bijection, a class that represents every assignment
$k \mapsto h(\cdot, k)$ through the embeddings $c_k$ also represents every
assignment through $c_{s(k)}$. The minimum over the class is therefore the same
for every $s$.
\end{proof}

\noindent\textbf{Lemma~\ref{lem:conflict} (Embedding conflict).}
\emph{Let the noise-level embeddings $c_1, \dots, c_B$ lie on a line, consecutive embeddings $\delta$
apart, and let noise level $s(i)$ be read through $c_i$. Let $F_k$ denote the
posterior mean of noise level $k$ in the model's output space, and let $f$ be
$\ell$-Lipschitz in its embedding in $L^2(\mu)$ for an input distribution
$\mu$. With $d_i = \lVert F_{s(i+1)} - F_{s(i)} \rVert_{L^2(\mu)}$ and
$e_k = \lVert f(\cdot, c_{s^{-1}(k)}) - F_k \rVert_{L^2(\mu)}$,
\[
\sum_{k} e_k^2 \;\ge\; \frac{1}{4} \sum_{i=1}^{B-1} \big(d_i - \ell\,\delta\big)_+^2 .
\]}
\begin{proof}
All norms are in $L^2(\mu)$. For consecutive positions $i$ and $i+1$ the
triangle inequality gives
\[
d_i \le \lVert F_{s(i+1)} - f(\cdot, c_{i+1}) \rVert + \lVert f(\cdot,
c_{i+1}) - f(\cdot, c_i) \rVert + \lVert f(\cdot, c_i) - F_{s(i)} \rVert
\le e_{s(i+1)} + \ell\,\delta + e_{s(i)} .
\]
Hence $e_{s(i)} + e_{s(i+1)} \ge (d_i - \ell\delta)_+$, and
$e_{s(i)}^2 + e_{s(i+1)}^2 \ge \tfrac12 (e_{s(i)} + e_{s(i+1)})^2 \ge
\tfrac12 (d_i - \ell\delta)_+^2$. Summing over $i = 1, \dots, B-1$ counts
each $e_k^2$ at most twice, so $2\sum_k e_k^2 \ge \tfrac12 \sum_i (d_i -
\ell\delta)_+^2$.
\end{proof}

\section{The mixture of Gaussian experiments}
\label{app:mog}

The data are $x_0 = \mu_c + \eta$ in 64 dimensions, with $c$ uniform on 64 components. We draw the means once in a random 16-dimensional subspace and transform them there to have covariance $\mathrm{diag}(\lambda)$ over the components. The first 10 entries of $\lambda$ are the top 10 eigenvalues of the CIFAR-10 training-pixel covariance on the $[-1, 1]$ scale, from 221.45 down to 10.95. The last 6 are 10, 9, 8, 7, 6 and 5. The noise $\eta$ is Gaussian with covariance $V\,\mathrm{diag}(s)\,V^\top$, where $V$ is orthonormal and its first 16 columns span the means. The entries of $s$ are 1 along the 16 directions of the means. Along the other 48 they are the eigenvalues of ranks 17 to 3072 of the same CIFAR-10 covariance, averaged over 48 consecutive blocks of 63 or 64 ranks, from 1.99 down to $9.6 \times 10^{-6}$.

We write every vector in the coordinates of $V$. For $y = x_0 + \sqrt{a}\,\varepsilon$ the posterior over the components has weights $w_c(y) \propto \exp\{\sum_i (y_i \mu_{ci} - \mu_{ci}^2/2)/(s_i + a)\}$. Given $c$, $x_0$ is Gaussian with mean $G\,y + (I - G)\,\mu_c$ and covariance $S$, where $G = \mathrm{diag}(s_i/(s_i + a))$ and $S = \mathrm{diag}(s_i a/(s_i + a))$. The posterior mean is therefore $\hat x(y) = G\,y + (I - G) \sum_c w_c(y)\,\mu_c$. The posterior covariance is $S + (I - G)\,B_w\,(I - G)$, with $B_w$ the covariance of the means under the weights. The law of a transition from $a$ to $b$ is the mixture, with weights $w_c(y)$, of the Gaussians with means $r\,y + (1-r)(G\,y + (I - G)\,\mu_c)$ and covariance $\tau I + (1-r)^2 S$.

The grid contains 128 transitions, spaced geometrically in $\sigma$ from 80 to 0.002. We estimate energy distance with a U-statistic on 4k samples from each side, drawn from the sampler's 5k outputs and from 20k data draws at $\sigma = 0.002$. Two independent draws from the data distribution give an estimated energy distance within 0.01 of zero.

In Sections~\ref{sec:order} and~\ref{sec:experiments}, each network fits $\hat x$ on its assigned transitions using the preconditioning of \citet{karras2022edm}, with $\sigma_{\mathrm{data}}^2$ equal to the data variance per coordinate. The regression target is $(x_0 - c_{\mathrm{skip}}\,y)/c_{\mathrm{out}}$. Each network is a multilayer perceptron with 3 hidden layers of the listed width and SiLU activations. Its input combines $c_{\mathrm{in}}\,y$ with the sines and cosines of $\log\sigma$ at 8 frequencies spaced geometrically from 0.25 to 8. Each network is trained with Adam for 80k steps at a batch size of 1024, using fresh pairs whose transitions are sampled uniformly from those assigned to the network. The one-cycle learning rate peaks at $2\times10^{-3}$. We evaluate excess error at each transition on 2k fresh pairs, measuring $\mathbb{E}\lVert (\hat x_{\mathrm{fit}} - \hat x)/c_{\mathrm{out}} \rVert^2$ per coordinate.

The composed sampler takes 128 steps, using the posterior mean from the network assigned to each transition. It starts from the marginal of its training pairs at $\sigma = 80$ and is evaluated on 5k samples. We report means over 3 seeds.

We compare these errors with three references. The closed-form linear denoiser is the posterior mean under a Gaussian with the data's mean and covariance; it has excess error 5.7 ($\times 100$) and its sampler has energy distance 0.13. One network of width 512 has excess error 0.54 and energy distance 0.03. Using the exact posterior mean at every transition gives energy distance 0.010 under DDPM and 0.007 under DDIM, at the level of two independent draws from the data distribution.

At a fixed total parameter count, the separate networks of Figure~\ref{fig:mogpar}b have widths 112, 239 and 494 for $K=4$ and 44, 104 and 230 for $K=16$, between 0.2\% and 1.4\% fewer parameters than the one network.

To test replacement in the setting of Theorem~\ref{prop:tes}, we retrain 1 of 16 separate networks of width 128 with a fresh seed, leaving the other 15 networks unchanged. The energy distance of the composed sampler changes by a small fraction of its variation across sampling seeds (Table~\ref{tab:mogcomp}).

\begin{table}[t]
\centering\small
\caption{The mixture of Gaussian experiments. Energy distance of the composed 128-step sampler to the data under the DDPM step: 16 separate networks of width 128, and the same with one network retrained at a fresh seed, mean $\pm$ standard deviation over 3 sampling seeds.}
\label{tab:mogcomp}
\begin{tabular}{lc}
\toprule
Composed model & energy distance \\
\midrule
16 separate networks & 0.067\,$\pm$\,0.009 \\
16 separate networks, 1 retrained & 0.065\,$\pm$\,0.009 \\
\bottomrule
\end{tabular}
\end{table}

For Section~\ref{sec:order}, a single network of the architecture described above fits all 128 transitions. At transition $k$, we replace the scalar $\log\sigma$ supplied to its sinusoidal noise-level embedding with the value at transition $s(k)$. The assignment $s$ either preserves the noise-level order, reverses it, shuffles within consecutive windows of $w$ transitions, or shuffles the full grid. For the learned embedding, each transition has its own 16 trainable entries, initialised at random. We also train a network without an embedding. All conditions keep the true noise level in the preconditioning, and conditions with the same seed share their initialisation and training pairs. We measure variation along the order of the embeddings by $\sum_i d_i^2$, computing $d_i$ exactly on the pooled preconditioned inputs from every transition. The grid of 32 transitions spans the same $\sigma$ range. Results are means over 5 seeds.

For the embedding comparison, we keep the networks, grid and training fixed and change only the noise-level embedding (Table~\ref{tab:embedding}). The learned embedding uses a trainable 16-dimensional vector per transition, while the random embedding uses a vector fixed at initialisation. At each seed the three embeddings share their network initialisation and training pairs.

\begin{table}[t]
\centering\small
\caption{CIFAR-10, one network of width 32 trained for 60k steps. Checkmarks identify
the noise-level embedding variants and the second training run. `Per transition' means
one embedding value per transition. FID uses 50k samples;
held-out denoising loss is averaged over 128 transitions.}
\label{tab:order}
\begin{tabular}{cccccccc}
\toprule
per transition & reversed & shuffled & learned & second training & & FID & loss per transition \\
\midrule
$\times$ & $\times$ & $\times$ & $\times$ & $\times$ & & 9.46 & 0.4643 \\
\checkmark & $\times$ & $\times$ & $\times$ & $\times$ & & 9.11 & 0.4640 \\
\checkmark & $\times$ & $\times$ & $\times$ & \checkmark & & 9.28 & 0.4642 \\
\checkmark & \checkmark & $\times$ & $\times$ & $\times$ & & 9.57 & 0.4633 \\
\checkmark & \checkmark & $\times$ & $\times$ & \checkmark & & 9.26 & 0.4638 \\
\checkmark & $\times$ & \checkmark & $\times$ & $\times$ & & 9.25 & 0.4761 \\
\checkmark & $\times$ & $\times$ & \checkmark & $\times$ & & 9.16 & 0.4654 \\
\bottomrule
\end{tabular}
\end{table}

The CIFAR-10 comparison in Table~\ref{tab:order} changes only the noise-level embeddings supplied to the network. The training pairs, their transition assignments, the input and output scalings, and the loss weight all retain the true noise level.

We repeat the comparison between ordered and shuffled noise-level embeddings at widths 64 and 128 (Table~\ref{tab:orderwidth}). At widths 64 and 128, 85\% and 86\% of the increase in loss occur in the noisiest quarter of the transitions, where loss is 5.1\% higher.

\section{Protocols}
\label{app:protocols}

Every CIFAR-10 network is the U-Net of \citet{karras2022edm} in its DDPM++ configuration (channel multipliers 2, 2, 2, 4 blocks per resolution, attention at $16\times16$, dropout 0.1), conditioned on the class label, with the width of each table as its base channel count. It trains for 60k steps at a batch size of 512 with Adam at learning rate $10^{-3}$, ramped up linearly over the first 10M images, and we evaluate the exponential moving average of its weights with a half-life of 500k images. The noise level is drawn from the log-normal distribution of \citet{karras2022edm} over the network's own range of $\sigma$, and the preconditioning and loss weight are theirs. Every CIFAR-10 evaluation takes one stochastic step per transition, 128 unless a table states otherwise, starting from $\sigma = 80$, and draws 5k samples per class. Random horizontal flips are the only augmentation.

On ImageNet-256, every model trains for 400k steps with a batch size of 256 on sd-vae-ft-ema latents, with random horizontal flips. Each evaluation takes 249 stochastic steps starting from transition 1. For the model with 4 separate networks, the networks' assigned ranges have boundaries at transitions 1, 63, 125, 187 and 250. Labels and noise are drawn on the CPU by sample index. We use no guidance and evaluate 50k samples against the ADM reference batch.

The boundaries of Table~\ref{tab:cuts} come from the movement of the exact posterior mean. For transition $k$, we draw inputs $y = x_0 + \sigma_k\varepsilon$ at its noisier level, compute the exact posterior means at $\sigma_k$ and $\sigma_{k+1}$ in the model's output space under the preconditioning of \citet{karras2022edm}, and take the root mean square difference per coordinate. The mixture uses 4,000 draws per transition. CIFAR-10 uses 2,048, and its posterior is exact for the empirical measure of the 50,000 training images, without the class label. The distances of Table~\ref{tab:perlevel} and the variation in the caption of Table~\ref{tab:cifarorder} use the class-conditional posterior means, which the class-conditional networks fit. Range $i$ ends where the cumulative movement first reaches $(i+1)/4$ of the total. The mixture's ranges hold 31, 19, 24 and 54 transitions from the noisiest, and CIFAR-10's hold 47, 28, 26 and 27. On CIFAR-10, the range edges in $\sigma$ lie at the geometric midpoints between consecutive levels. Table~\ref{tab:cutsfid} gives the FID of the composed CIFAR-10 models.

\begin{table}[t]
\centering\small
\caption{CIFAR-10 at 14M, FID over 50k samples on one evaluator, of the four separate networks of Table~\ref{tab:cuts} under each set of boundaries.}
\label{tab:cutsfid}
\begin{tabular}{lc}
\toprule
Boundaries & FID \\
\midrule
equal split & 7.11 \\
own posterior mean & 7.50 \\
other data set's posterior mean & 7.09 \\
\bottomrule
\end{tabular}
\end{table}

\section{Retraining and Replacing Networks}
\label{app:modularity}\label{sec:replacement}

Once the ranges share no parameters, the following bound relates the change of replaced transition laws to the output distribution.

\begin{theorem}[Modularity]
\label{prop:tes}
Let the transitions carry disjoint parameters and let the sampler draw
$y_{k+1} \sim \hat p_k(\cdot \mid y_k)$ for $k = 0, \dots, B-1$ from
$y_0 \sim \mathcal{N}(0, a_0 I)$. A second sampler from the same prior
that replaces $\hat p_j$ by $\hat p'_j$ for every $j$ in a set $J$ of
transitions satisfies
\[
\mathrm{TV}\big(\mathrm{law}(y_B),\ \mathrm{law}(y'_B)\big)
\;\le\; \sum_{j \in J} \mathbb{E}\,\mathrm{TV}\big(\hat p_j(\cdot \mid
y_j),\ \hat p'_j(\cdot \mid y_j)\big),
\]
where the expectation is taken under the marginal at transition $j$ of
the chain with all replacements before $j$ already applied.
\end{theorem}

Replacing every fitted law by its forward-process law gives the following bound.

\begin{corollary}[Error of the composed model]
\label{cor:tes}
With $J$ every transition, $\hat p'_j$ the transition laws $p_j$ of the
forward process and the prior $p_{a_0}$, Theorem~\ref{prop:tes}
gives $\mathrm{TV}(\mathrm{law}(y_B), p_{\mathrm{data}} \ast
\mathcal{N}(0, a_B I)) \le \mathrm{TV}(\mathcal{N}(0, a_0 I), p_{a_0})
+ \sum_k \mathbb{E}\,\mathrm{TV}(\hat p_k, p_k)$, whose sum vanishes with exact laws at every transition count and whose first term vanishes as $a_0$ grows.
\end{corollary}

For the 4 architectures of Section~\ref{sec:division}, almost all of the excess held-out denoising loss is confined to the transformer's transitions, with little excess at the boundaries between architectures (Table~\ref{tab:hetero}).

\section{Ablations on the partition}
\label{app:ablations}

A shared network uses the same width at every noise level, and increasing that width lowers FID (Table~\ref{tab:capacity}).

Separate networks allow different network widths at different noise levels. We divide the noise grid into two halves and train one network on each half, using a network of width 64 for the noisier half and a network of width 40 for the cleaner half. We repeat the experiment with the widths exchanged. This comparison keeps the total parameter count fixed and tests which half benefits more from the wider network. Assigning the wider network to the cleaner half lowers FID from 5.07 to 4.95. When both networks have width 64, the total parameter count increases and FID falls further, to 4.23 (Table~\ref{tab:capacity}).

\begin{table}[t]
\centering\small
\caption{CIFAR-10 FID over 50k samples on one evaluator. The upper block varies the width of a single network. In the lower block, one network fits each half of the noise grid. The $64{:}40$ and $40{:}64$ allocations exchange the wider network between the noisier and cleaner halves at the same total parameter count. The $64{:}64$ model has a larger total parameter count. Widths are listed as noisier half:cleaner half.}
\label{tab:capacity}
\begin{tabular}{lcccc}
\toprule
1 network, width & 24 & 32 & 40 & 64 \\
\midrule
FID & 12.69 & 9.45 & 7.49 & 4.68 \\
\midrule
Network widths, noisier half\,:\,cleaner half & $64{:}40$ & $40{:}64$ & & \\
\midrule
FID & 5.07 & 4.95 & & \\
\midrule
Same width for both networks & $64{:}64$ & & & \\
\midrule
FID & 4.23 & & & \\
\bottomrule
\end{tabular}
\end{table}

\section{Adjacency}
\label{app:adjacency}

Dividing the transitions among networks can narrow each network, create boundaries and remove parameter sharing across ranges. These changes have different signatures across the noise grid.

If the gap came from the boundaries between networks, the loss would rise near them, and the increase would grow with the number of separate networks.

Interference through shared parameters predicts a different change: separating the noise levels should reduce the loss within each network's range. Theorem~\ref{thm:noorder} makes the unrestricted local fitting problems independent; the measurements below test whether finite networks incur any boundary cost once width is accounted for.

Dividing contiguous ranges among separate networks also narrows the noise range each network must fit. On the mixture experiments, we control for this by distributing successive transitions among the $K$ separate networks in turn. Each network receives every $K$-th transition and spans the full noise range. For 4 separate networks their error then matches that of a single network of the same width to within 2\% at widths 256 and 512 (Figure~\ref{fig:mogpar}a). This control separates the effect of independent training from that of a shorter range, which lowers the error at the same width.

For CIFAR-10, we compare held-out denoising loss for 4 separate networks of width 64 and 16 separate networks of width 32. Relative to a single network of the same width, the loss beside a boundary is slightly higher for 4 separate networks and slightly lower for 16 (Table~\ref{tab:seams}). Away from the boundaries the loss is lower by a few per cent, several times the boundary differences. Both boundary differences are nevertheless larger than the difference between two training runs of one network, so they are resolved. The reduction in interior loss has the sign predicted by interference through shared parameters.

The 16 separate networks of width 32 fit the held-out transitions better than the 4 separate networks inside the ranges and no worse at their boundaries (Table~\ref{tab:f1}). At widths 16 and 32, the model with 16 separate networks has FID 0.5 to 0.7 higher than the model with 4 (Table~\ref{tab:splitting}), the same size as the spread over 3 training runs of one network at width 32 (FID 9.42, 9.45 and 9.96).

At both 14M and 56M, a separate network's loss relative to a single network of the same width is highest at the end of its range nearest transition 78, where the last range starts (Figure~\ref{fig:profile}). For the ranges covering transitions 0 to 57, 58 to 67 and 68 to 77, relative loss rises toward the clean end: the increase from the first quarter to the last is 0.4 to 1.7 points at 56M. The range covering transitions 78 to 127 has its highest relative loss at the noisy end, with $-1.0\%$ in its first quarter and $-6.4\%$ in its last. The mixture experiments show the same pattern around the peak of the single network's error, near transition 35 (Figure~\ref{fig:orderenters}c). There, every range has its highest relative error at the end nearest that peak, and 3 of the 4 ranges lie on its clean side (Figure~\ref{fig:mogpar}c).

\begin{table}[t]
\centering\small
\caption{CIFAR-10, held-out denoising loss for 4 and 16 separate networks of width 32, averaged over the listed groups of transitions. We use the training loss weight and 10k held-out pairs per transition with fixed noise. The boundaries of the 4 separate networks are also boundaries of the 16. The last column counts the transitions at which the 16 separate networks have lower loss.}
\label{tab:f1}
\begin{tabular}{lcccc}
\toprule
Transitions & count & 4 separate networks & 16 separate networks & 16 lower on \\
\midrule
all & 128 & 0.4530 & 0.4439 & 100 \\
interior to both & 98 & 0.5431 & 0.5313 & 83 \\
boundary of the 16 only & 24 & 0.1643 & 0.1643 & 13 \\
boundary of both & 6 & 0.1348 & 0.1340 & 4 \\
\bottomrule
\end{tabular}
\end{table}

\begin{table}[t]
\centering\small
\caption{CIFAR-10, 4 separate networks with different architectures compared with 4 width-64 networks at the same total parameter count. Excess is the increase in held-out denoising loss for the mixed-architecture model relative to the width-64 model, averaged over each network's transitions. Share is each network's contribution to the excess summed over all 128 transitions. Transitions are numbered from the noisiest. The last row averages the 6 transitions beside boundaries between networks, which are also included in the network rows. Both compositions use the same width-64 network.}
\label{tab:hetero}
\begin{tabular}{lcccc}
\toprule
Network & parameters & transitions & excess & share \\
\midrule
SiT, depth 12, width 256 & 14.5M & 78--127 & $+3.9\%$ & 97.8\% \\
U-Net, width 96, 1 block per resolution & 14.0M & 68--77 & $+1.3\%$ & 1.7\% \\
U-Net, width 32, 6 blocks per resolution & 14.1M & 58--67 & $+0.5\%$ & 0.5\% \\
U-Net, width 64 & 14.0M & 0--57 & $0.0\%$ & 0.0\% \\
\midrule
6 transitions beside a boundary & & & $+0.7\%$ & 0.6\% \\
\bottomrule
\end{tabular}
\end{table}

\section{Where a Full Shuffle Raises the Error}
\label{app:perlevel}

Table~\ref{tab:perlevel} gives, level by level, where a full shuffle raises the error. Across noise levels, the rise correlates with the ordered network's error on the mixture at every width and on CIFAR-10. On CIFAR-10 it also correlates with how much the inputs of a level overlap those of its ordered neighbours, which measures how little the noisy input says about its own noise level. The rise has near-zero or negative correlation with the change in the distances of Lemma~\ref{lem:conflict} on both data sets and with input overlap on the mixture.

\begin{table}[t]
\centering\small
\caption{Where a full shuffle raises the error, level by level. Spearman correlation across the noise levels between the rise and, at each level: the distances of Lemma~\ref{lem:conflict} to the levels whose embeddings sit next to its own after the shuffle, minus those to its ordered neighbours; the error of the ordered network; and the overlap of its inputs with those of its ordered neighbours. The last column gives the largest share of the rise in one quarter of the grid, with quarters counted from the noisiest. Mixture: excess error, means over 5 seeds. CIFAR-10: held-out denoising loss of one network of width 32.}
\label{tab:perlevel}
\begin{tabular}{llcccc}
\toprule
Data & width & distances & ordered error & input overlap & share of the rise \\
\midrule
mixture & 256 & $-0.02$ & 0.35 & $-0.25$ & 0.37, second quarter \\
mixture & 512 & $-0.20$ & 0.70 & $-0.17$ & 0.49, second quarter \\
mixture & 1024 & $-0.13$ & 0.69 & $-0.08$ & 0.65, second quarter \\
CIFAR-10 & 32 & 0.00 & 0.83 & 0.65 & 0.72, first quarter \\
\bottomrule
\end{tabular}
\end{table}

We also space a scalar embedding of $\log\sigma$ by the distances of Lemma~\ref{lem:conflict} (Table~\ref{tab:geomcode}). The scalar code keeps the range of $\log\sigma$, and each step covers an equal share of the distance between consecutive posterior means. The code stretches the cleanest quarter of the grid, where the distances on the pooled inputs are largest, and compresses the rest. The error rises in the stretched quarter and falls elsewhere. On average it rises at widths 256 and 512 and falls at width 1024.

\begin{table}[t]
\centering\small
\caption{Mixture, one network over 128 transitions with the scalar embedding of $\log\sigma$ in noise-level order or spaced so that each step covers an equal share of the distances of Lemma~\ref{lem:conflict} between consecutive posterior means, over the same range. Mean excess error per transition, $\times 100$, over 5 seeds; the number of seeds at which the spaced embedding has the lower error; and the change of the spaced embedding's error in each quarter of the grid, from the noisiest.}
\label{tab:geomcode}
\begin{tabular}{lcccrrrr}
\toprule
Width & ordered & spaced & lower on & \multicolumn{4}{c}{change by quarter} \\
\midrule
256 & 1.72 & 1.78 & 1 of 5 & $-0.1\%$ & $-1.0\%$ & $-1.0\%$ & $+169.5\%$ \\
512 & 0.56 & 0.61 & 0 of 5 & $-3.7\%$ & $-3.6\%$ & $-12.1\%$ & $+650.5\%$ \\
1024 & 0.37 & 0.34 & 4 of 5 & $-21.5\%$ & $-15.4\%$ & $-44.2\%$ & $+186.9\%$ \\
\bottomrule
\end{tabular}
\end{table}

We also evaluate each network at noise levels on which it was never trained. Between every 2 consecutive transitions we take the level at the geometric mean of their noise levels and give the network the midpoint of their 2 embedding values; the preconditioning uses the true noise variance of that level, and we compute the excess error from the exact posterior mean at that level. With the embeddings in order or reversed, the error at these 127 unseen levels equals the error at the trained levels at every width (Table~\ref{tab:unseen}). After a full shuffle the 2 embeddings next to an unseen level belong to distant noise levels, and the error there is about 50 to several hundred times the error at the trained levels. When 4 or 16 blocks are moved, the unseen levels inside a block have the same error as the trained levels, and the error increases at the 3 or 15 levels where 2 moved blocks meet. A network therefore generalises to unseen noise levels through the levels adjacent to them in the embedding, in either direction.

\begin{table}[t]
\centering\footnotesize\setlength{\tabcolsep}{3pt}
\caption{Mixture, one network over 128 transitions, evaluated at its trained levels and at the 127 unseen levels between consecutive transitions, each through the midpoint of the 2 transitions' embedding values: mean excess error $\pm$ standard deviation, $\times 100$, over 5 seeds, on the networks of Table~\ref{tab:blocks}. For the moved blocks, the unseen levels are divided into those where 2 moved blocks meet and the rest.}
\label{tab:unseen}
\begin{tabular}{llccccccc}
\toprule
 & & & & \multicolumn{2}{c}{4 blocks} & \multicolumn{2}{c}{16 blocks} & \\
\cmidrule(lr){5-6}\cmidrule(lr){7-8}
Width & levels & ordered & reversed & inside & between blocks & inside & between blocks & full shuffle \\
\midrule
256 & trained & 1.72\,{\scriptsize$\pm$0.10} & 1.69\,{\scriptsize$\pm$0.08} & \multicolumn{2}{c}{1.77\,{\scriptsize$\pm$0.07}} & \multicolumn{2}{c}{1.96\,{\scriptsize$\pm$0.07}} & 2.35\,{\scriptsize$\pm$0.29} \\
 & unseen & 1.73\,{\scriptsize$\pm$0.10} & 1.70\,{\scriptsize$\pm$0.08} & 1.74\,{\scriptsize$\pm$0.07} & 191\,{\scriptsize$\pm$76} & 1.94\,{\scriptsize$\pm$0.08} & 1699\,{\scriptsize$\pm$720} & 111\,{\scriptsize$\pm$120} \\
\midrule
512 & trained & 0.56\,{\scriptsize$\pm$0.01} & 0.55\,{\scriptsize$\pm$0.02} & \multicolumn{2}{c}{0.59\,{\scriptsize$\pm$0.01}} & \multicolumn{2}{c}{0.64\,{\scriptsize$\pm$0.02}} & 0.80\,{\scriptsize$\pm$0.04} \\
 & unseen & 0.57\,{\scriptsize$\pm$0.01} & 0.55\,{\scriptsize$\pm$0.02} & 0.58\,{\scriptsize$\pm$0.01} & 450\,{\scriptsize$\pm$76} & 0.65\,{\scriptsize$\pm$0.02} & 717\,{\scriptsize$\pm$76} & 56\,{\scriptsize$\pm$8} \\
\midrule
1024 & trained & 0.37\,{\scriptsize$\pm$0.02} & 0.34\,{\scriptsize$\pm$0.01} & \multicolumn{2}{c}{0.43\,{\scriptsize$\pm$0.07}} & \multicolumn{2}{c}{0.34\,{\scriptsize$\pm$0.08}} & 0.69\,{\scriptsize$\pm$0.24} \\
 & unseen & 0.37\,{\scriptsize$\pm$0.02} & 0.34\,{\scriptsize$\pm$0.01} & 0.42\,{\scriptsize$\pm$0.06} & 1047\,{\scriptsize$\pm$1700} & 0.36\,{\scriptsize$\pm$0.08} & 8337\,{\scriptsize$\pm$17000} & 297\,{\scriptsize$\pm$130} \\
\bottomrule
\end{tabular}
\end{table}

\section{Further Related Work}
\label{app:related}

\subsection{From a physical process to a family of marginals}
\label{sec:lineage}

Diffusion models began as a physical process. Using a Markov chain motivated by non-equilibrium thermodynamics, \citet{sohl2015deep} gradually corrupted the data and learned a reverse chain that restored it. \citet{song2019generative} estimated the score of the data at a sequence of noise levels and sampled with annealed Langevin dynamics. In \citet{ho2020ddpm}, each training step draws one noise level and one noisy sample from the closed-form marginal at that level, without running the forward chain. \citet{song2021sde} then joined both models in a stochastic differential equation in continuous time, reversed by a second such equation or by a probability-flow ordinary differential equation. This step strengthened the physical description, and the common view of a diffusion model as one continuous function of the noise level comes from it.

Later work changed the forward process, the noise schedule, and the path between data and noise. \citet{song2021ddim} observed that the objective depends on the forward process only through its marginals, and a model trained with the DDPM objective serves a family of non-Markovian processes with the same marginals. \citet{kingma2021vdm} showed that in continuous time the variational bound depends on the noise schedule only through the signal-to-noise ratio at its two ends, which leaves the schedule between them free. \citet{karras2022edm} expressed every formulation through the noise level $\sigma$, with the time variable as one parameterisation of it, and chose the distribution of noise levels in training separately from the steps taken at sampling. The noise need not be Gaussian: deterministic degradations such as blurring and masking train generators of the same form \citep{bansal2023cold}, and electrostatics can replace the physics of diffusion \citep{xu2022pfgm}. Flow matching, rectified flow and stochastic interpolants need no stochastic differential equation. They keep a path of marginals and regress the velocity that transports samples along it \citep{lipman2023flow, liu2023rectified, albergo2023stochastic}. The coupling between data and noise along the path is also free, and minibatch optimal-transport couplings straighten the paths \citep{tong2024improving, pooladian2023multisample}. Generator Matching extends the construction to any Markov process through its generator \citep{holderrieth2025generator}. All of these formulations keep one network for all noise levels.

\subsection{The object fitted at one noise level}
\label{sec:onelevel}

Tweedie's formula gives the posterior mean of the clean sample in terms of the score of the noisy marginal \citep{robbins1956empirical, efron2011tweedie}. Differentiating this mean with respect to the noisy state and multiplying by the noise variance gives the posterior covariance. \citet{song2021ddim} observed that the objective depends on the forward process only through its marginal at each noise level, leaving both the coupling between consecutive levels and the levels visited at sampling free to be chosen. They also noted that, when the parameters are not shared across noise levels, the optimum does not depend on the loss weights, since each term is optimised on its own, as \citet{song2019generative} had noted for their conical combination of per-level objectives. They used this separability to justify the unweighted DDPM loss as a surrogate and to show that their non-Markovian objectives share its optimum. We use the same separability to establish invariance to relabelling (Theorem~\ref{thm:noorder}) and then test how parameter sharing affects the fitting error (Sections~\ref{sec:experiments} and~\ref{sec:order}).

Existing samplers also choose the order of noise levels after training. The stochastic sampler of \citet{karras2022edm} and Restart sampling \citep{xu2023restart} return to noisier levels during a run. DDIM uses a single network across all noise levels, connecting their predictions through shared parameters (Section~\ref{sec:axis}). We remove that connection between groups of levels and measure its effect on fitting error and sample quality.

\subsection{One parameter set across the noise levels}

Fast samplers change the numerical solver \citep{song2021ddim, lu2022dpm}. Another line of work models large transitions with richer families of finite laws \citep{xiao2022ddgan, shaul2025tm}, and discrete-state diffusion replaces the Gaussian with masking or other discrete corruptions \citep{sahoo2024mdlm, lou2024sedd}. We fit groups of transition laws with separate networks and compose them at sampling.

\subsection{Objectives that couple the noise levels}
\label{sec:coupledobj}

Distillation and consistency training couple several levels. Progressive distillation trains a student to match two teacher steps with one \citep{salimans2022progressive}. Consistency models require a network's outputs at different points along a trajectory of the probability-flow ODE to match one another \citep{song2023cm}. Consistency trajectory models learn maps between any two levels on that trajectory \citep{kim2024ctm}. These objectives add terms involving two or more noise levels ($\mathcal{M}$ in Appendix~\ref{app:laws}) and therefore put into the objective an order that their starting diffusion objective does not contain.

\subsection{Divisions that keep the noise level whole}

Parallel training distributes the data, parameters or optimiser states across machines \citep{shoeybi2019megatron, huang2019gpipe, rajbhandari2020zero}, synchronising every training step and retaining shared parameters across noise levels. Other methods weight or group noise levels to reduce interference through shared parameters \citep{hang2023minsnr, go2023negative, park2024dtr}. We remove parameter sharing between groups of noise levels.

Expert denoisers assign different intervals of the noise schedule to different networks \citep{balaji2022ediff}. These experts branch from a shared backbone after joint training. \citet{lee2024meme} assign networks with different architectures to disjoint intervals and train each network on its own interval with a set probability and on the whole process otherwise. In their control without this mixing, the experts train independently from initialisation on their own intervals and yield an FID 0.7 to 0.8 points higher on FFHQ. This control uses our construction, and we examine what it reveals about the object that training fits. In our experiments, the cost of independent training comes from each network's smaller width: it shrinks as the total parameter count grows, and is almost absent at boundaries between networks (Section~\ref{sec:experiments}).

Decentralized diffusion models divide the data in place of the noise levels \citep{mcallister2025ddm}. Each expert trains on one cluster of the data in full isolation, the ensemble optimises the same objective as a single model trained on all the data, and a learned router mixes the experts' predictions at every sampling step. A heterogeneous variant trains some experts with the DDPM objective and others with flow matching, converts their predictions to one parameterisation at sampling, and in one analysis hands sampling from one kind of expert to the other at a fixed noise level \citep{jiang2026heterogeneous}. Their construction shows that the data can be divided; ours shows that the noise levels can. A division along the noise levels needs no router, since each sampling step is supplied by the one network that holds its transition. By Lemma~\ref{lem:pairing}, switching the sampler from one expert to another trained on a different forward process still reaches the data distribution when the state arrives with the marginal the next expert was trained on.

Conditional computation uses a learned router to select interchangeable experts for each input \citep{shazeer2017moe}. In our model, the noise level determines which network supplies a transition. Cascaded stages and language-model experts train independently along other axes \citep{ho2022cascaded, li2022btm}; the latter are combined by ensembling their outputs or averaging their parameters. We compose transition laws in noise-level order, and Theorem~\ref{prop:tes} bounds the effect of replacing any law.

\end{document}

%% file: math_commands.tex
\usepackage{amsmath,amsfonts,bm}

\def\eqref#1{equation~\ref{#1}}

\def\1{\bm{1}}

\DeclareMathAlphabet{\mathsfit}{\encodingdefault}{\sfdefault}{m}{sl}
\SetMathAlphabet{\mathsfit}{bold}{\encodingdefault}{\sfdefault}{bx}{n}

